\documentclass{article} 
\usepackage{iclr2017_conference,times}
\usepackage{hyperref}
\usepackage{url}
\usepackage{graphicx}
\usepackage{color}
\usepackage{subfigure}
\usepackage{amssymb}
\usepackage{listings}

\graphicspath{figs}

\title{Diet Networks: Thin Parameters for\\Fat Genomics}

\author{
Adriana Romero\thanks{Equal contribution.}, Pierre Luc Carrier\footnotemark[1],\\ \textbf{Akram Erraqabi, Tristan Sylvain,} \\ \textbf{Alex Auvolat, Etienne Dejoie}  \\
Montreal Institute for Learning Algorithms \\
Montreal, Quebec, Canada \\
\texttt{firstName.lastName}@umontreal.ca, except\\
\texttt{adriana.romero.soriano@umontreal.ca} \\
and \texttt{pierre-luc.carrier@umontreal.ca}
\And
Marc-Andr\'{e} Legault$^1$, Marie-Pierre Dub\'{e}$^{1,2,3}$\\
$^1$University of Montreal, Faculty of Medicine \\
$^2$Montreal Heart Institute,\\
$^3$Beaulieu-Saucier Pharmacogenomics Centre\\
Montreal, Quebec, Canada \\
\texttt{marc-andre.legault.1@umontreal.ca}\\
\texttt{marie-pierre.dube@umontreal.ca}
\And
Julie G. Hussin \\
Wellcome Trust Centre for Human Genetics\\ University of Oxford \\
Oxford, UK \\
\texttt{julieh@well.ox.ac.uk}
\And
Yoshua Bengio \\
Montreal Institute for Learning Algorithms \\
Montreal, Quebec, Canada \\
\texttt{yoshua.umontreal@gmail.com}
}

\iclrfinalcopy 

\begin{document}

\maketitle

\begin{abstract}
Learning tasks such as those involving genomic data often poses a serious challenge: the number of input features can be orders of magnitude larger than the number of training examples, making it difficult to avoid overfitting, even when using the known regularization techniques. We focus here on tasks in which the input is a description of the genetic variation specific to a patient, the single nucleotide polymorphisms (SNPs), yielding millions of ternary inputs. Improving the ability of deep learning to handle such datasets could have an important impact in medical research, more specifically in precision medicine, where high-dimensional data regarding a particular patient is used to make predictions of interest. Even though the amount of data for such tasks is increasing, this mismatch between the number of examples and the number of inputs remains a concern. Naive implementations of classifier neural networks involve a huge number of free parameters in their first layer (number of input features times number of hidden units): each input feature is associated with as many parameters as there are hidden units. We propose a novel neural network parametrization which considerably reduces the number of free parameters. It is based on the idea that we can first learn or provide a distributed representation for each input feature (e.g. for each position in the genome where variations are observed in data), and then learn (with another neural network called the parameter prediction network) how to map a feature's distributed representation (based on the feature's identity not its value) to the vector of parameters specific to that feature in the classifier neural network (the weights which link the value of the feature to each of the hidden units). This approach views the problem of producing the parameters associated with each feature as a multi-task learning problem. We show experimentally on a population stratification task of interest to medical studies that the proposed approach can significantly reduce both the number of parameters and the error rate of the classifier.
\end{abstract}

\section{Introduction}
\label{sec:intro}


Medical datasets often involve a dire imbalance between the number of training examples and the number of input features, especially when genomic information is used as input to the trained predictor. This is problematic in the context where we want to apply deep learning (which typically involves large models) to precision medicine, i.e., making patient-specific predictions using a potentially large set of input features to better characterize the patient. This paper proposes a novel approach, called Diet Networks, to reparametrize neural networks to considerably reduce their number of free parameters when the input is very high-dimensional and orders of magnitude larger than the number of training examples.

Genomics is the study of the genetic code encapsulated as DNA in all living organisms' cells. Genomes contain the instructions to produce and regulate all the functional components needed to guide the development and adaptation of living organisms. In the last decades, advances in genomic technologies resulted in an explosion of available data, making it more interesting to apply advanced machine learning techniques such as deep learning.
%
Learning tasks involving genomic data and already tackled by deep learning include: using Convolutional Neural Networks (CNNs) to learn the functional activity of DNA sequences (Basset package, \cite{Kelley028399}, predicting effects of noncoding DNA (DeepSEA, \cite{Zhou2015}), investigating the regulatory role of RNA binding proteins in alternative splicing \citep{alipanahi2015predicting}, inferring gene expression patterns \citep{Chen2016, DeepChrome} and population genetic parameters \citep{Sheehan2016} among others (see \cite{leung2016} for a detailed example). Noticeably, most of these techniques are based on sequence data where convolutional or recurrent networks are appropriate. When the full DNA sequence is unavailable, such as when data is acquired through genotyping, other methods need to be used. All this work shows that deep learning can be used to tackle genomic-related tasks, paving the road towards a better understanding of the biological impact of DNA variation.

Applying deep learning to human genetic variation holds the promise of identifying individuals at risk for medical conditions. Modern genotyping technologies usually target millions of simple variants across the genome, called single nucleotide polymorphisms (SNPs). These genetic mutations result from substitutions from one nucleotide to another (eg. A to C), where both versions exist within a population. In modern studies, as many as 5 millions SNPs can be acquired for every participant. These datasets differ from other types of genomic data because they focus on the genetic differences between individuals which represents a space of high dimensionality where sequence-context information is unavailable.
In medical genetics, these variants are tested for their association with a trait of interest, an approach termed genome-wide association study (GWAS). This methodology aims at finding genetic variants implicated in disease susceptibility, etiology and treatment. 

An important confounding factor in GWAS is population stratification, which arises because both disease prevalence and genetic profiles vary from one population to the other. Although most GWAS have been restricted to homogeneous populations, dimensionality reduction techniques are generally used to account for population-level genetic differences \citep{eigenstrat}. Our experiments compare such dimensionality reduction techniques (based on principal components analysis, PCA) to the proposed Diet Network parametrization, as well as with standard deep networks.

Recently, several machine learning methods have been successfully applied to detect population stratification, based on the presence of systematic differences in genetic variation between populations. For instance, Support Vector Machines (SVM) models have been used multiple times to infer recent genetic ancestry of sub-continental populations (\cite{Haasl2013}), and local ancestry in admixed populations (SupportMix, \cite{Omberg2012}, 23andMe, Inc.). However SVM methods are very sensitive to the the kernel choice and the parameters. They also tend to overfit the model selection criterion which usually induces a limitation in its predictive power. 

In this work, we are interested in predicting the genetic ancestry of an individual from their SNP data using a novel deep learning approach, Diet Networks, which allow us to considerably reduce the number of free parameters. Therefore, we propose to tackle this problem by introducing a multi-task architecture in which the problem of predicting the appropriate parameters for each input feature is considered like a task in itself, and the same {\em parameter prediction network} is used for all of the hundreds of thousands of input features. This parameter prediction network learns to predict these feature-specific parameters as a function of a distributed representation of the feature identity, or feature embedding. The feature embedding can be learned as part of end-to-end training or using other datasets or a priori knowledge about the features. What is important is that two features which are similar in some appropriate sense (in terms of their interactions with other features or other variables observed in any dataset) end up having similar embeddings, and thus a similar parameter vector as output of the parameter prediction network. A practical advantage of this approach is that the parameter prediction network can generalize to new features for which there is no labeled training data (without the target to be predicted by the classifier), so long as it is possible to derive an embedding for that feature (for example using just the unlabeled observations of co-occurences of that feature with other features in human genomes).

An interesting consideration is that {\em from the point of the parameter prediction network, each feature is an example: more features now allow to better train the parameter prediction network.} It is like if we were considering not the data matrix itself but {\em its transpose}. This is actually how the Diet Network implementation processes the data, by using the transpose of the matrix of input values as the input part of the learning task for the parameter prediction network.

The idea of having two networks interacting with each other and with one producing parameters for the other is well rooted in the machine learning literature \citep{Bengio+al-1991,Schmidhuber1992,Gomez05,Stanley09,denil2013predicting,Andrychowicz2016}.
Recent efforts in the same direction include works such as \citep{Bertinetto16,DeBrabandere16, Ha16} that use a network to predict the parameters of a Convolutional Neural Network (CNN). \citet{DeBrabandere16} introduce a dynamic filter module that generates network filters conditioned on an input. \citet{Bertinetto16} propose to learn the parameters of a deep model in one shot, by training a second network to predict the parameters of the first from a single exemplar. Hypernetworks \citep{Ha16} explore the idea of using a small network to predict the parameters of another network, training them in an end-to-end fashion. The small network takes as input the feature embedding from the previous layer and learns the parameters of the current layer. 

To the best of our knowledge, deep learning has never been used so far to tackle the problem of ancestry prediction based on SNP data. Compared to other approaches that attempt to learn model parameters using a parameter prediction network, our main goal is to reduce the large number of parameters required by the model, by considering the input features themselves as sub-tasks in a multi-task view of the learning problem, as opposed to constructing a model with even higher capacity, as seen, e.g. in \citep{Ha16}. Our approach is thus based on building an embedding of these tasks (the features) in order to further reduce the number of parameters.

We evaluate our method on a publicly available dataset for ancestry prediction, the 1000 Genomes dataset\footnote{http://www.internationalgenome.org/}, that best represents the human population diversity. Because population-specific differences in disease and drug response are widespread, identifying an individual's ancestry heritage based on SNP data is a very important task to help detect biological causation and achieve good predictive performance in precision medicine. Most importantly, ancestry-aware approaches in precision genomics will reduce the hidden risks of genetic testing, by preventing spurious diagnosis and ineffective treatment.


\section{Method}
\label{sec:method}

In this section, we describe the Diet Networks as well as the feature embeddings used by the model.

\subsection{Model}
\label{ssec:model}
\begin{figure}[ht!]
\centering
\scalebox{1.2}{
\subfigure[]{\includegraphics[width=0.185\textwidth]{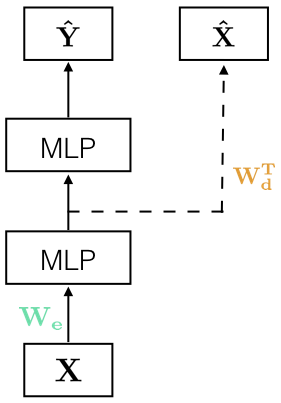}\label{fig:MLPbasic}}\hfill\hspace{1cm}
\subfigure[]{\includegraphics[width=0.09\textwidth]{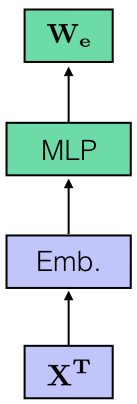}\label{fig:MLPenc}}\hfill\hspace{1cm}
\subfigure[]{\includegraphics[width=0.09\textwidth]{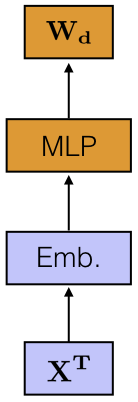}\label{fig:MLPdec}}\hfill}
\caption{Our model is composed of 3 networks, one basic and two auxiliary networks: (a) a basic discriminative network with optional reconstruction path (dashed arrow), (b) a network that predicts the input fat layer parameters, and (c) a network that predicts the reconstruction fat layer parameters (if any). First layer in the ''prediction networks" (b, c) represents embedding (Emb.). Each MLP block may contain any number of hidden layers. $\mathbf{W_e}$ and $\mathbf{W_d^T}$ represent the parameters of the fat hidden layer and the fat reconstruction layer of the basic network (a), respectively. These parameters are predicted by auxiliary networks (b) and (c) -- also called parameter prediction networks -- to reduce the number of free parameters of (a).}
\label{fig:method}
\end{figure}

Our model aims at reducing the number of free parameters that a network trained on \emph{fat data} would typically have.

Let $\mathbf{X} \in \mathbb{R}^{N\times N_d}$ be a matrix of data, with $N$ samples and $N_d$ features, where $N \ll N_d$ (e. g. $N$ being approximately 100 times smaller than $N_d$).  We build a multi-layer perceptron (MLP), which takes $\mathbf{X}$ as input, computes a hidden representation and outputs a prediction $\mathbf{\hat{Y}}$. Optionally, the MLP may generate a reconstruction $\mathbf{\hat{X}}$ of the input data from the hidden representation. Figure \ref{fig:MLPbasic} illustrates this basic network architecture. Let $\mathbf{x_i}$ be one data sample, i.e. a row in $\mathbf{X}$. The standard formulations to compute its hidden representation $\mathbf{h_i}$, output prediction $\mathbf{\hat{y}_i}$ and reconstruction $\mathbf{\hat{x}_i}$ are given by
\begin{equation}
\mathbf{h_i} = f(\mathbf{x_i}), \quad
\mathbf{\hat{y_i}} = g(\mathbf{h_i}), \quad
\mathbf{\hat{x_i}} = r(\mathbf{h_i}),
\end{equation}
where $f$, $g$ and $r$ are non-linear functions.

The number of parameters of the first hidden layer of the architecture grows linearly with the dimensionality of the input data:
\begin{equation}
\mathbf{h_{i}^{(1)}} = f_1(\mathbf{x_i}\mathbf{W_e} + \mathbf{b_e}),
\end{equation}
where $\mathbf{W_e}$ and $\mathbf{b_e}$ are the layer's parameters. Using \emph{fat data} such as the one described in Section \ref{sec:intro}, leads to a parameter explosion in this layer, hereafter referred to as \emph{fat hidden layer}. To give the reader an intuition, consider the case of having an input with $N_d = 300K$, and a hidden layer with $N_h^1 = 100$, the number of parameters of such a layer would be 30M. The same happens to the number of parameters of the optional reconstruction layer, hereafter referred to as \emph{fat reconstruction layer}.  

In order to mitigate this effect, we introduce an auxiliary network to predict the fat layers' parameters. The auxiliary network takes as input the {\em transposed} data matrix $\mathbf{X^T}$, extracts a feature embedding and learns a function of this embedding, to be used as parameters of a fat layer: 
\begin{equation}
\mathbf{\left(W_e\right)_{j:}} = \phi(\mathbf{e_j}),
\end{equation}
where $\mathbf{e_j}$ represents the embedding of a feature in $\mathbf{X^T}$, $\phi$ is a non-linear function and $\mathbf{\left(W_e\right)_{j:}}$ is the $j$-th row of $\mathbf{W_e}$.
This means that each feature is associated with the vector of values it takes in the dataset (e.g. across the patients). Other representations could be used, e.g., derived from other datasets in which those features interact. Figure \ref{fig:MLPenc} shows a prediction network which is an auxiliary network that predicts the parameters of the fat hidden layer of our basic network. Following the same spirit, Figure \ref{fig:MLPdec} highlights the interaction between a second prediction network that predicts the fat reconstruction layer parameters and the basic network. The architectures of both auxiliary networks may share the initial feature embedding.

The feature embeddings used in the auxiliary networks allow us to substantially reduce the number of free parameters of the fat layers of the basic architecture. The auxiliary network should predict a matrix of weights of size $N_d \times N_h^1$ from a feature embedding. Consider a feature embedding that would transform each $N$-dimensional feature into a $N_f$-dimensional vector, where $N_f < N$. The auxiliary network would learn a function $\phi:\mathcal{R}^{N_f} \rightarrow \mathbb{R}^{N_h^1}$. Thus, the fat hidden layer of our basic architecture would have $N_f \times N_h^1$ free parameters (assuming a single layer MLP in the auxiliary network), instead of $N_d \times N_h^1$. Following our previous example, where $N_d = 300K$ and $N_h^1 = 100$, using an auxiliary network with previously-obtained feature embeddings of dimensionality $N_f = 500$ would reduce the number of free parameters of the basic network by a factor of 600 (from 30M to 50K).

The model is trained end-to-end by minimizing the following objective function
\begin{equation}
\mathcal{H}(\mathbf{\hat{Y}}, \mathbf{Y}) + \gamma ||\mathbf{\hat{X}} - \mathbf{X}||_2^2,
\end{equation}
where $\mathcal{H}$ refers to the cross-entropy, $\mathbf{Y}$ to the true classification labels and $\gamma$ is a tunable parameter to balance the supervised and the reconstruction losses.

\subsection{Feature embeddings}
\label{ssec:embeddings}
The feature embeddings used by the auxiliary networks can be either pre-computed or learnt offline, as well as learnt jointly with the rest of the architecture. In theory, any kind of embedding could be used, as long as we keep in mind that the goal is to reduce the number of free parameters of the basic model. In this work, we considered random projections \citep{Bingham01}, histograms (which are akin to bag-of-words representations), feature embeddings learnt offline \citep{Mikolov-et-al-NIPS2013} and feature embeddings jointly learnt with the rest of the proposed architecture. 

\textbf{Random projection:}
Randomly initializing an MLP defines a random projection. By using such a projection to encode the high-dimensional feature space into a more manageable lower-dimensional space, we were able to obtain decent results.

\textbf{Per class histogram:}
For a given SNP, we can define a histogram of the values it can take over the whole population. Once normalized, this yields 3 values per SNP, corresponding to the proportion of the population having the values 0, 1 and 2 respectively for that SNP. After initial tests showed this was too coarse a representation for the dataset, we instead chose to consider the per-class proportion of the three values. With 26 classes in the 1000 Genomes dataset, this yields an embedding of size 78 for each feature. By this method, the matrix $\mathbf{X^T}$ is summarized as a $N_d \times 78$ matrix, where $N_d$ is the number of SNPs in the dataset.

\textbf{SNPtoVec:}
In \cite{Mikolov-et-al-NIPS2013}, the authors propose a word embedding that allows good reconstruction of the words' context (surrounding words) by a neural network. SNPs do not have a similarly well-defined positional context (SNPs close together in our ordering might very well be independent) so our embedding is instead built by training a denoising autoencoder (DAE) \citep{vincent:icml08} on the matrix $\mathbf{X}$. Thus, the DAE learns to recover the values of missing SNPs by leveraging their similarities and cooccurences with other SNPs. Once the DAE is trained, we obtain an encoding for each feature by feeding to the DAE an input where only that feature is active (the other features are set to 0s) and computing the hidden representation of the autoencoder for that single-feature input.


\textbf{Embedding learnt end-to-end from raw data:} In this case, we consider the feature embedding to be another MLP, whose input corresponds to the values that a SNP takes for each of the training samples and, whose parameters are learnt jointly with the rest of the network. Note that the layer(s) corresponding to the feature embedding are shared among auxiliary networks. For experiments reported in Section \ref{sec:exp}, we used a single hidden layer as embedding.

\section{Data: The 1000 Genomes Project}
\label{ssec:1000gen}

\begin{figure}[ht!]
\centering
\subfigure[]{\includegraphics[width=0.45\textwidth]{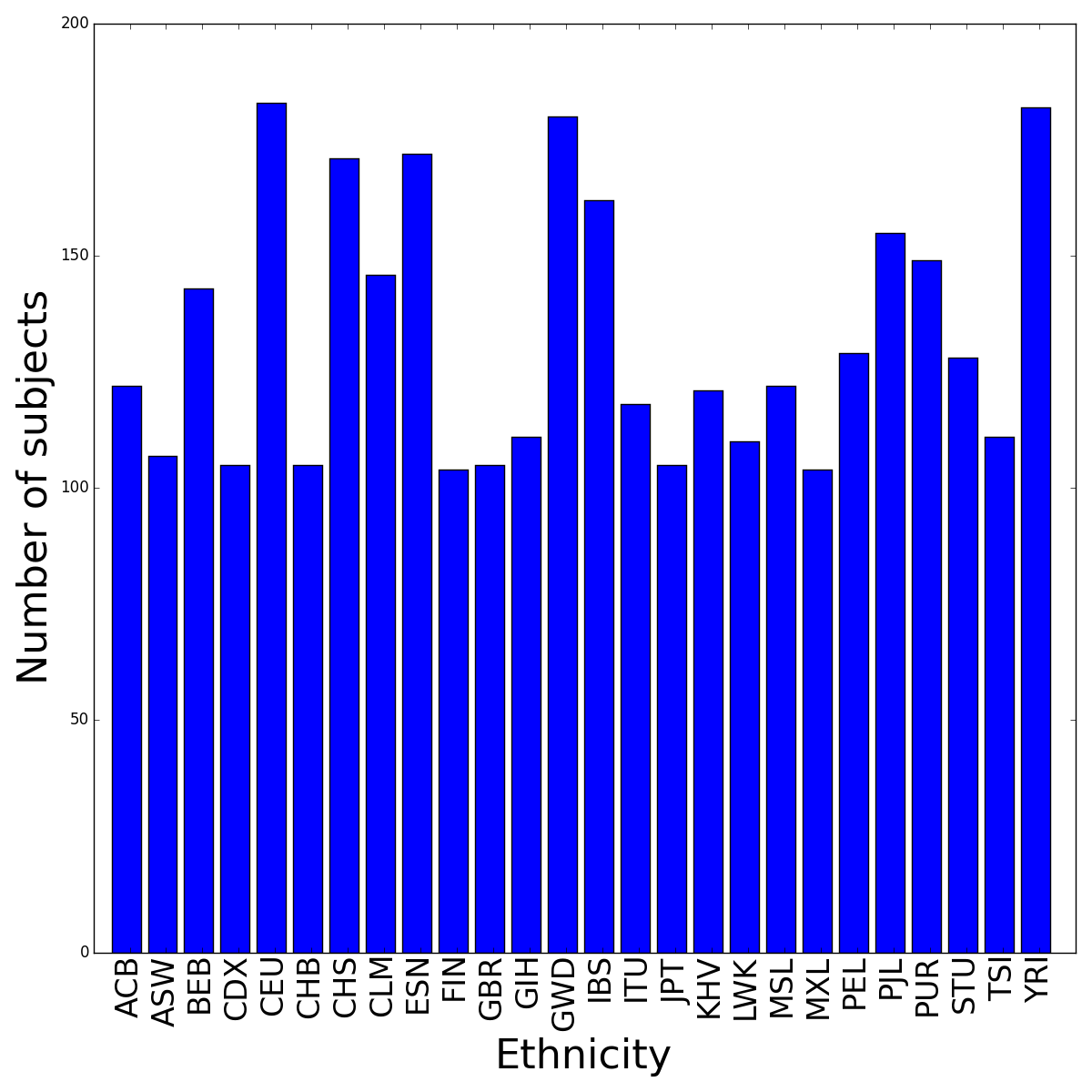}\label{fig:eth_histo}}\hfill
\subfigure[]{\includegraphics[width=0.45\textwidth]{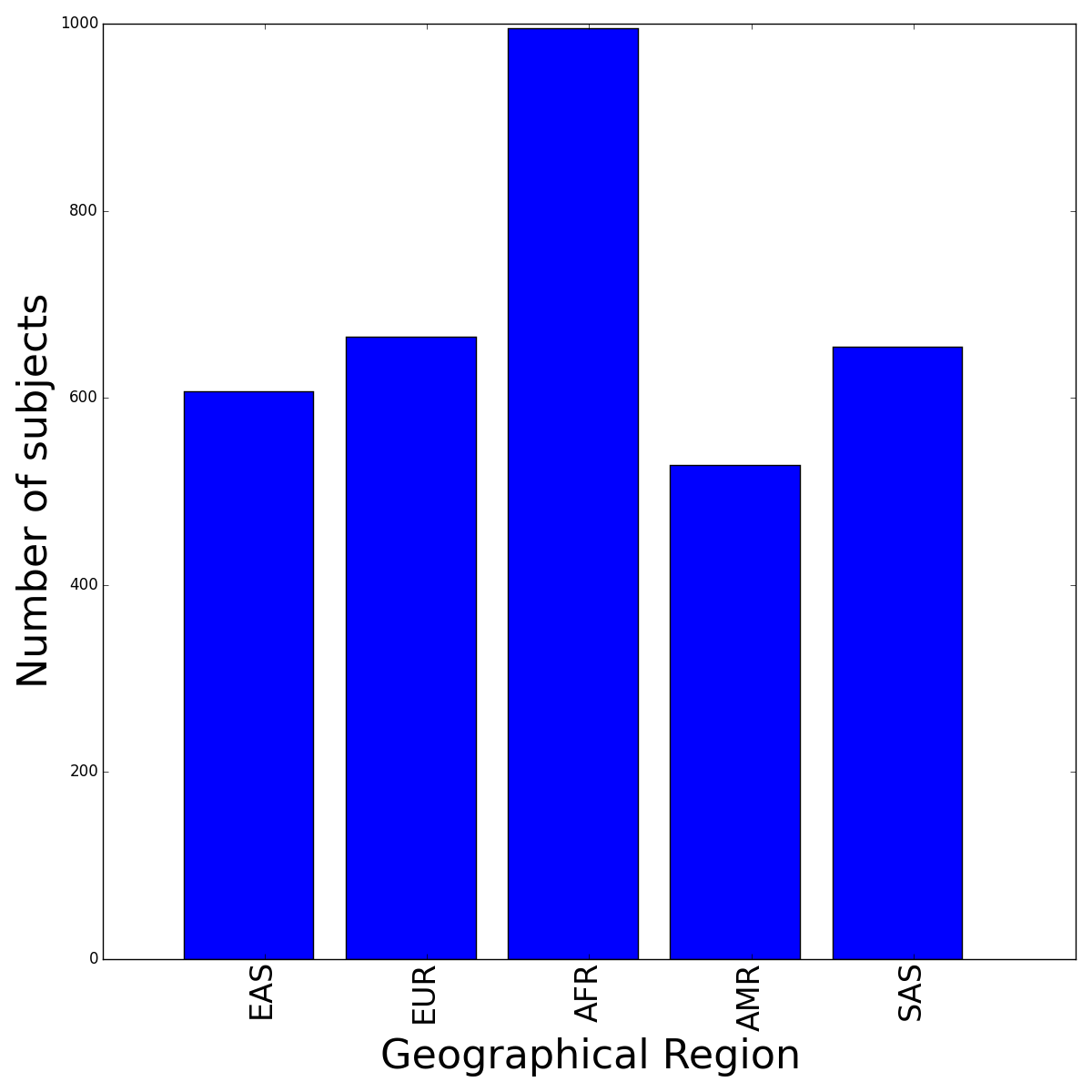}\label{fig:cont_histo}}\hfill
\caption{\textbf{The 1000 Genomes population distribution:}(a) Ethnicity; (b) Geographical Region.}
\label{fig:histos}
\end{figure}

The 1000 Genomes project is the first project to sequence the genomes of a large number of people in populations worldwide, yielding the largest public catalog of human genetic variants to date \cite{1000Gpaper}. This allowed large-scale comparison of DNA sequences from populations, thanks to the presence of genetic variation. Individuals of the 1000 Genomes project are samples taken from 26 populations over the world, which are grouped into 5 geographical regions. Figure \ref{fig:eth_histo} shows a histogram derived from the 1000 Genomes data, depicting the frequency of individuals per population (ethnicity). Analogously, Figure \ref{fig:cont_histo} depicts the frequency of individuals per geographical region.

In this dataset, we included 315,345 genetic variants with frequencies of at least 5\% in 3,450 individuals sampled worldwide from 26 populations, interrogated using microarray genotyping technology: the Genome-Wide Human SNP Array 6.0 by Affymetrix. The mutated state is established by comparison to the Genome Reference Consortium human genome (build 37). Since individuals have 2 copies of each genomic position, a sampled individual can have 0, 1 or 2 copies of a genetic mutation, hereafter referred to as an individual genotype. We excluded SNPs positioned on the sex chromosomes and only included SNPs in approximate linkage equilibrium with each other, such that genotypes at neighboring positions are only weakly correlated ($r^{2} < 0.5$).

\section{Experiments}
\label{sec:exp}

In this section, we describe the model architectures, and report and discuss the obtained results.

\begin{figure}[ht!]
\centering
\subfigure[]{\includegraphics[width=0.5\textwidth]{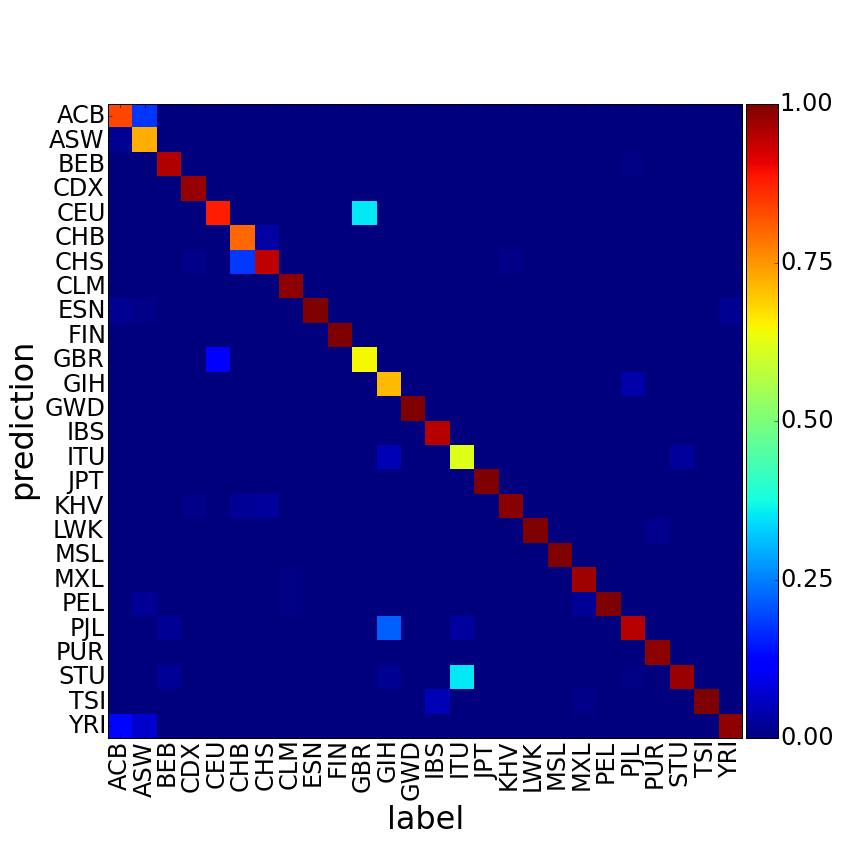}\label{fig:eth_cm}}\hfill
\subfigure[]{\includegraphics[width=0.5\textwidth]{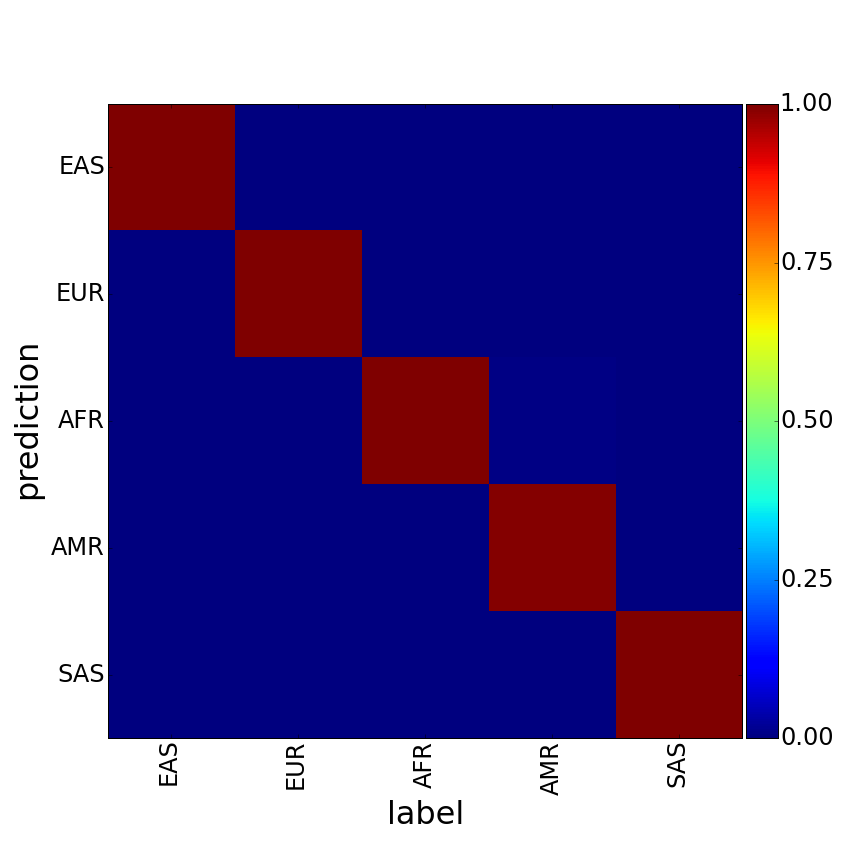}\label{fig:cont_cm}}\hfill
\caption{\textbf{Results of our best model:} (a) Confusion matrix per ethnicity; (b) Confusion matrix per large geographical region. The 1000 Genomes legend for population abbreviations can be found in the appendix.}
\label{fig:cm}
\end{figure} 

\subsection{Model Architecture} 
We experimented with simple models both in the auxiliary networks and the basic architecture, which yielded very promising results. We designed a basic architecture with 2 hidden layers followed by a softmax layer to perform ancestry prediction. We trained this architecture with and without the assistance of the auxiliary network. Similarly, the auxiliary networks were build by stacking a hidden layer on top of one of the feature embeddings described in Section \ref{ssec:embeddings}. In the reported experiments, all hidden layers have 100 units. All models were trained by means of stochastic gradient descent with adaptive learning rate \citep{Tieleman12}, both for $\gamma=0$ and $\gamma=10$, using dropout, limiting the norm of the weights to 1 and/or applying weight decay to reduce overfitting.\footnote{The code to reproduce the experiments can be found here: https://github.com/adri-romsor/DietNetworks}.

\subsection{Results}
Given the relatively small amount of samples in the 1000 Genomes data, we report results obtained by 5-fold cross validation of the model. We split the data into 5 folds of equal size. A single fold is retained for test, whereas three of the remaining folds are used as training data and the final fold is used as validation data. We repeated the process 5 times (one per fold) and report the means and standard deviations of results on the different test sets.

Table \ref{tab:1000_genomes} summarizes the results obtained for each model. First, we observe that, for most of Diet Network architectures, training with an reconstruction term in the loss ($\gamma > 0$) reduces the misclassification error and provides a lower standard deviation over the folds, suggesting more robustness to variations in the learnt feature embedding.

Training the models end-to-end, with no pre-computed feature embedding, yielded higher misclassification error than simply training the basic model, which could be attributed to the fact that adding the prediction networks makes a difficult, high-dimensional optimization problem even harder. As a general trend, adding pre-computed feature embeddings achieved better performance (lower error), while allowing to significantly reduce the number of free parameters in the fat layers of the model. Among the tested feature embeddings, random projections achieved good results, highlighting the potential of the model when reducing the number of free parameters.

Using the SNP2Vec embedding, trained to exploit the similarities and co-occurences between the SNPs, in conjunction with the Diet Networks framework obtains slightly better results than the model using a random projection. The addition of the reconstruction criterion does not appear to reduce the number of errors made by the model but it does appear to reduce the variance of the results, as observed on the other models.

Despite its simplicity, the per class histogram encoding (when used with a reconstruction criterion) yielded the best results.
Note that this encoding is the one with the fewest number of free parameters in the fat layers, with a reduction factor of almost 4000 w.r.t. the analogous basic model (with reconstruction). Figure \ref{fig:eth_cm} shows the mean results obtained with the histogram embedding. As shown in the figure, when considering the ethnicity, the main misclassifications involve ethnicities likely to display very close genetic proximity, such as British from England and Scotland, and Utah residents with Northern and Western ancestry (likely to be immigrants from England), or Indian Telugu and Sri Lankan Tamil for instance. However, the model achieves almost 100\% accuracy when considering the 5 geographical regions. 

We also compared the performance of our model to the principal component analysis (PCA) approach, commonly used in the genomics domain, to select subgroups of individuals in order to perform more homogeneous analysis. The number of principal components (PCs) is chosen according to their significance, and usually varies from one dataset to another, being 10 the \emph{de facto} standard for small datasets. However, in the case of the 1000 Genomes dataset, we could go up to 50 PCs. Therefore, we trained a linear classifier on top of PCA features, considering 10 and 50 PCs, 100 PCs to match the number of feature used in the other experiments, as well as 200 PCs. Using 200 PCs yielded better performance, but going beyond that saturated in terms of misclassification error (see Section\ref{app:pca} in Appendix for more details). Adding hidden layers to the classifier didn't help either (see reported results for several MLP configurations before the linear classifier).

\begin{table}
\centering
\footnotesize
 \begin{tabular}{|c || c | c |} 
 \hline
 \textbf{Model \& Embedding} & \textbf{Mean Misclassif. Error. (\%)} & \textbf{\# of free parameters}\\
 \hline
 Basic & $8.31 \pm 1.83$ &31.5M\\ \hline
 Raw end2end & $8.88 \pm 1.42$ &217.2k \\ \hline
 Random Projection & $9.03 \pm 1.20$ &10.1k \\ \hline
 SNP2Vec & $\mathbf{7.60 \pm 1.28}$ &$\mathbf{10.1k}$ \\ \hline
 Per class histograms & $7.88 \pm 1.40$ &7.9k\\ \hline \hline
 Basic with reconstruction & $7.76 \pm 1.38$ &63M \\ \hline
 Raw end2end with reconstruction & $8.28 \pm 1.92$ &227.3k\\ \hline
 Random Projection with reconstruction & $8.03 \pm 1.03$ &20.2k \\ \hline
 SNP2Vec with reconstruction & $7.88 \pm 0.72$ &20.2k \\ \hline
 Per class histograms with reconstruction & $\mathbf{7.44 \pm 0.45}$ & $\mathbf{15.8k}$\\ \hline \hline
 \textbf{Traditional approaches} & \multicolumn{2}{|c|}{\textbf{Mean Misclassif. Error. (\%)}} \\ \hline
 PCA (10 PCs) & \multicolumn{2}{|c|}{$20.56 \pm 3.20$}\\ \hline
 PCA (50 PCs) & \multicolumn{2}{|c|}{$12.29 \pm 0.89$}\\ \hline
PCA (100 PCs) & \multicolumn{2}{|c|}{$10.52 \pm 0.25$}\\ \hline
PCA (200 PCs) & \multicolumn{2}{|c|}{$9.33 \pm 1.24$}\\ \hline
PCA (100 PCs) + MLP(50)& \multicolumn{2}{|c|}{$12.67 \pm 0.67$}\\ \hline
PCA (100 PCs) +  MLP(100)& \multicolumn{2}{|c|}{$12.18 \pm 1.75$}\\ \hline
PCA (100 PCs) + MLP(100, 100)& \multicolumn{2}{|c|}{$11.95 \pm 2.29$}\\ \hline
 \end{tabular}
 \caption{Results for 1000 Genomes ancestry prediction. Raw end2end, random projection and SNP2Vec embeddings have dimensionality 100, whereas per class histograms has dimensionality 78. Note that the reported number of free parameters corresponds to the free parameters of the fat layers of the models.}
 \label{tab:1000_genomes}
\end{table}

\section{Conclusion}
\label{sec:concl}
In this paper, we proposed Diet Networks, a novel network parametrization, which considerably reduces the number of free parameters in the fat layers of a model when the input is very high dimensional. We showed how using the parameter prediction networks, yielded better generalization in terms of misclassification error.  Notably, when using pre-computed feature embeddings that maximally reduced the number of free parameters, we were able to obtain our best results. We validated our approach on the publicly available 1000 genomes dataset, addressing the relevant task of ancestry prediction based on SNP data. This work demonstrated the potential of deep learning models to tackle domain-specific tasks where there is a mismatch between the number of samples and their high dimensionality. 

Given the high accuracy achieved in the ancestry prediction task, we believe that deep learning techniques can improve standard practices in the analysis of human polymorphism data. We expect that these techniques will allow us to tackle the more challenging problem of conducting genetic association studies. Hence, we expect to further develop our method to conduct population-aware analyses of SNP data in disease cohorts. The increased power of deep learning methods to identify the genetic basis of common diseases could lead to better patient risk prediction and will improve our overall understanding of disease etiology.




\subsubsection*{Acknowledgments}
The authors would like to thank the developers of Theano \cite{Theano-2016short} and Lasagne \cite{lasagne}. We acknowledge the support of the following agencies for research funding and computing support: Imagia, CIFAR, Canada Research Chairs, Compute Canada and Calcul Qu\'{e}bec. J.G.H. is an EPAC/Linacre Junior Research Fellow funded by the Human Frontiers Program (LT-001017/2013-L). Special thanks to Val\`{e}ria Romero-Soriano, Xavier Grau-Bov\'{e} and Margaux Luck for their patience sharing genomic biology expertise; as well as to Michal Drozdzal, Caglar Gulcehre and Simon J\'{e}gou for useful discussions and support.

\bibliography{iclr2017_conference}
\bibliographystyle{iclr2017_conference}

\appendix

\section{The 1000 Genomes Project Legends}

\subsection{Population ethnicity legend} 
\textbf{ACB}: African Caribbeans in Barbados\\
\textbf{ASW}: Americans of African Ancestry in SW USA\\
\textbf{BEB}: Bengali from Bangladesh\\
\textbf{CDX}: Chinese Dai in Xishuangbanna\\
\textbf{CEU}: Utah Residents (CEPH) with Northern and Western Ancestry\\ 
\textbf{CHB}: Han Chinese in Bejing \\
\textbf{CHS}: Southern Han Chinese\\ 
\textbf{CLM}: Colombians from Medellin\\
\textbf{ESN}: Esan in Nigeria\\
\textbf{FIN}: Finnish in Finland\\ 
\textbf{GBR}: British in England and Scotland \\
\textbf{GIH}: Gujarati Indian from Houston\\
\textbf{GWD}: Gambian in Western Divisions in the Gambia\\
\textbf{IBS}: Iberian Population in Spain\\
\textbf{ITU}: Indian Telugu from the UK\\
\textbf{JPT}: Japanese in Tokyo\\
\textbf{KHV}: Kinh in Ho Chi Minh City\\
\textbf{LWK}: Luhya in Webuye\\
\textbf{MSL}: Mende in Sierra Leone\\
\textbf{MXL}: Mexican Ancestry from Los Angeles\\
\textbf{PEL}: Peruvians from Lima\\
\textbf{PJL}: Punjabi from Lahore\\
\textbf{PUR}: Puerto Ricans\\
\textbf{STU}: Sri Lankan Tamil from the UK\\
\textbf{TSI}: Toscani in Italia\\ 
\textbf{YRI}: Yoruba in Ibadan\\

\subsection{Geogprahical region legend} 

\textbf{AFR}: African\\
\textbf{AMR}: Ad Mixed American\\
\textbf{EAS}: East Asian\\
\textbf{EUR}: European\\
\textbf{SAS}: South Asian

\section{Obtaining the 1000 Genomes data} 

SNP data for the 1000G dataset was downloaded from \url{ftp://ftp.1000genomes.ebi.ac.uk:21/vol1/ftp/release/20130502/supporting/hd_genotype_chip/}
\begin{lstlisting}
- ALL.wgs.nhgri_coriell_affy_6.20140825.genotypes_has_ped.vcf.gz
\end{lstlisting} 
\begin{lstlisting}
- affy_samples.20141118.panel
\end{lstlisting}

\textbf{Representative commands:}
 
With PLINK v1.90b2n 64-bit \url{https://www.cog-genomics.org/plink2}

\begin{lstlisting}[language=bash,breaklines]
#convert vcf to plink format (bed), and only keeping common markers (minor allele frequency > 0.05 in combined sample)
> plink --vcf $path/ALL.wgs.nhgri_coriell_affy_6.20140825.genotypes_has_ped.vcf.gz --maf 0.05 --out $path/affy_6_biallelic_snps_maf005_aut --not-chr X Y MT --make-bed
 
#produce a pruned subset of markers that are in approximate linkage equilibrium with each other
> plink --bfile $path/affy_6_biallelic_snps_maf005_aut --indep-pairwise 50 5 0.5 --out $path/affy_6_biallelic_snps_maf005_aut
 
#exclude markers to get pruned subset                                                                                        
> plink --bfile $path/affy_6_biallelic_snps_maf005_aut --exclude $path/affy_6_biallelic_snps_maf005_aut.prune.out --recode A --out $path/affy_6_biallelic_snps_maf005_thinned_aut_A

\end{lstlisting}

For information on how to download this pre-processed dataset directly, please email Adriana Romero or Pierre Luc Carrier.

\section{PCA components and misclassification error}
\label{app:pca}
\begin{figure}[ht!]
\centering
\includegraphics[width=0.45\textwidth]{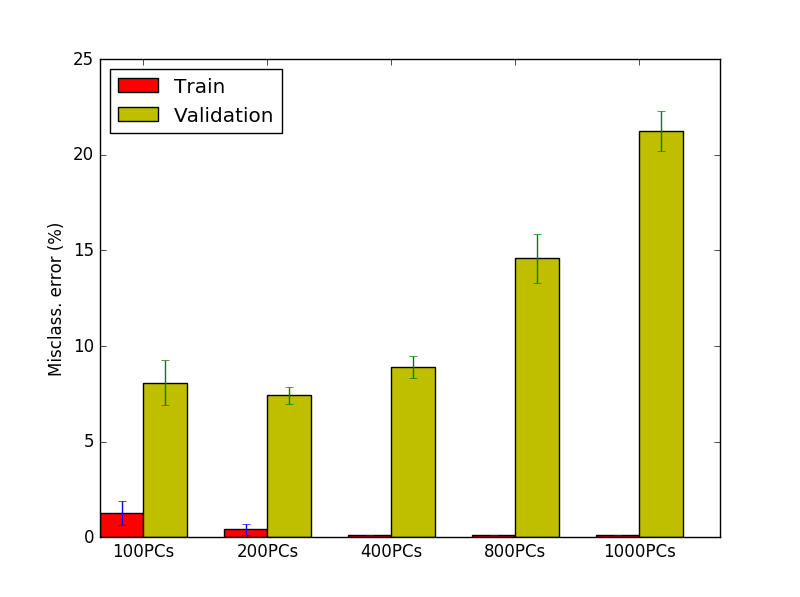}
\label{fig:pca}
\caption{\textbf{PCA:} Train and validation misclassification error for different numbers of PCs.}
\label{fig:pca_pcs}
\end{figure}

In this section, we analyze the influence of increasing the number of PCs used to perform classification. Figure \ref{fig:pca_pcs} depicts the obtained results when considering 100, 200, 400, 800 and 1000 PCs. As shown in the figure, the best validation error comes with PC200. Further increasing the number of PCs improves the training error but does not generalize well on the validation set. 


\end{document}